# Underwater Fish Detection using Deep Learning for Water Power Applications

Wenwei Xu
*Marine Sciences Division*
*Pacific Northwest National Laboratory*
Seattle, WA, USA
wenwei.xu@pnnl.gov

Shari Matzner
*Operational Systems Technology Division*
*Pacific Northwest National Laboratory*
Sequim, WA, USA
shari.matzner@pnnl.gov

*Abstract*— Clean energy from oceans and rivers is becoming a reality with the development of new technologies like tidal and instream turbines that generate electricity from naturally flowing water. These new technologies are being monitored for effects on fish and other wildlife using underwater video. Methods for automated analysis of underwater video are needed to lower the costs of analysis and improve accuracy. A deep learning model, YOLO, was trained to recognize fish in underwater video using three very different datasets recorded at real-world water power sites. Training and testing with examples from all three datasets resulted in a mean average precision (mAP) score of 0.5392. To test how well a model could generalize to new datasets, the model was trained using examples from only two of the datasets and then tested on examples from all three datasets. The resulting model could not recognize fish in the dataset that was not part of the training set. The mAP scores on the other two datasets that were included in the training set were higher than the scores achieved by the model trained on all three datasets. These results indicate that different methods are needed in order to produce a trained model that can generalize to new data sets such as those encountered in real world applications.

## I. Introduction

Underwater video is used for studying marine and riverine ecosystems, mainly for observing animal abundance and behavior in the context of environmental monitoring. Optical video provides detailed information in a form that is easy to interpret because humans naturally interpret their world visually. But human analysis of video takes time, so some level of automation is needed to process large video data sets in a timely, efficient manner in order to provide information needed for permitting and adaptive management decisions.

Computer vision and machine learning have proven effective for automating similar monitoring applications, such as video surveillance. Underwater imagery presents unique challenges that include sudden illumination changes, unequal spectral propagation (i.e., color content is affected by distance), low contrast, clutter in the form of floating vegetation, and changes in visibility due to turbidity [1].

Recently, a research tool for studying fish in underwater video was developed by the Fish4Knowledge project [2], and the dataset consisting of mostly coral reef fish were made publicly available. The web-based software provides researchers with a suite of computer vision methods that can be applied to underwater video for detecting, tracking and classifying fish around coral reefs [3]. The software, which includes modules for workflow management and archiving events in a database, provides a framework for offline post-processing of large video data sets.

There have been several other studies on underwater video fish detection using machine learning or deep learning methods, mostly focusing on coral reef fish species from public datasets like Fish4Knowledge. A few studies being reviewed include:

- Choi [4] first applied foreground detection method to extract fish object window and then used customized Convolutional Neural Networks (CNN) to classify 15 tropical fish species at LifeCLEF challenge. Choi's model achieve recall of near 0.9 and precision exceed 0.8.

- Li et al. [5] applied Fast R-CNN on ImageCLEF fish images to classify 12 coral fish species, and achieved mAP of 0.81.

- Zhang et al. [6] developed an unsupervised segmentation framework based on fish movement to automatically generate annotation labels, and showed promising results that auto-generated labels can be used to train fish detection models.

- Rathi et al. [7] combined Faster R-CNN with three classification networks (ZF Net, CNN-M, and VGG-16) to detect 50 fish and crustaceans species from Queensland beaches and estuaries. The regional proposal method consist of a regional proposal network coupled with a classifier network, and achieved 0.82 mAP.

- Mandal et al. [8] classified 21 tropical fish species in Fish4Knowledge dataset. A foreground enhancement preprocessing step is followed by a 3 layer CNN, and achieved 96% accuracy.

- Marini et al. [9] combined an image segmentation process, a k-fold Cross-Validation frame work feature selection process, and a binary classification process to determine whether fish exit in a segments and eventually count the fish blobs to evaluate the temporal fish abundance using data collected through marine cabled video-observatory.

However, most of the previous underwater fish optical analysis have focused on classification of coral fish in shallow waters or well-illuminated deep water where fish are abundant and visible. Few have studied detection of fish under more challenging underwater environment, the high turbidity, high velocity, murky water, especially around the development of marine and hydrokinetic (MHK) energy projects, or river hydropower projects, where the fish are often not colorful and hardly visible.

The few studies that are most similar to ours are: Kratzert and Mader [10, 11] developed FishNet to filter out fish migration videos at fish ladders that do not contain fish and used VGG to classify 6 fish species in Austrian rivers. At 2017 SeaCLEF challenge [12], one of our annotated dataset (Igiugig) was listed as a detection challenge to detect unclear objects (salmon) in underwater videos. Only one team submitted one run for the challenge, indicated the complexity of detection in low resolution underwater videos. Another analysis of the Igiugig dataset combined background subtraction and classification method and achieved promising results on semi-automatically detecting fish, however only used a subset of the videos [13]. To our knowledge, there have not been other applications of deep learning for MHK and hydropower projects yet.

Our contributions to the area of study are:

- Contributed 3 different sources MHK and hydropower underwater fish videos and manual annotations (Section II);
- Applied transfer learning and evaluated one of the state-of-the-art real time detection algorithms, YOLO [14], on challenging underwater environment video data (Section III and IV);
- Evaluated the feasibility of directly using our trained model to detect fish in another MHK or hydropower site (Section IV).

## II. MHK AND HYDROPOWER UNDERWATER FISH VIDEO DATASETS

Marine environments are challenging to study, and the growing development of marine and hydrokinetic (MHK) energy is driving the need for methods to observe underwater environments to monitor environmental effects. High-priority environmental risk uncertainties for MHK energy projects include the interaction of animals with energy devices, noise levels, and changes to marine animal distribution and habitat use at the space/time scales of device arrays [15]. These potential environmental risks pose challenges for permitting and licensing MHK projects, requiring novel and expensive approaches to monitor and evaluate potential interactions with organisms. Regulatory requirements, particularly related to endangered and threatened species and marine mammals, make near-continuous monitoring of the marine environment at MHK projects an essential part of quantifying and managing these risks. Continuous monitoring only becomes practical with automated event detection.

Optical images and video can provide valuable information for evaluating interactions between organisms and MHK energy converters [16], and can provide more detail than sonar data in energetic environments. The information contained in optical data includes aggregate measures of time-based occurrence, along with individual animal shape, color, relative position and movement direction. This information can be used to infer abundance, species distribution, behavior, and the likelihood of injury or mortality. The current practice for using optical data is to collect the data and then, post-collection, have human experts visually analyze the data and note events of interest.

However, extracting the relevant information from underwater imagery and video can be labor-intensive. In one study comparing video analysis to visual surveys, subject matter experts spent between 30 minutes to 1 hour, on average, reviewing each 4.5 minute transect video [17]. Oftentimes, the data is sampled so that, for example, only the first ten minutes of every hour is reviewed. This methodology is not practical for the long term monitoring needed to characterize temporal patterns of activity around MHK devices, nor effective for detecting rare but important events like the appearance of an endangered marine mammal. Automatic filtering is needed to reduce optical data to relevant events for expert review, shifting the effort from data management and processing to expert analysis and interpretation that informs decision-making.

Video data is needed to develop and validate algorithms. The data must contain examples of animals present in the scene and examples of "empty" video when no animals are present. A variety of data is needed --recorded in different locations, using different cameras and containing different species – in order to develop robust algorithms and to evaluate the software in a variety of conditions. Some of the data must be annotated with bounding boxes around objects of interest -- in this case, fish. This type of data, specific to MHK, is not readily available and is expensive to produce.

TABLE II. THREE HYDROKINETIC AND HYDROPOWER UNDERWATER VIDEO DATASETS

| Dataset | Voith Hydro | Wells Dam | Igiugig |
|---|---|---|---|
| Source | Aquatera | Douglas County Public Utility Dist. | Ocean Renewable Power Co. |
| Description | Recorded at EMEC in Fall of Warness, Scotland during the deployment of the Voith Hydro 1MW HyTide tidal energy converter. | Recorded through the fish passage viewing window at the dam in eastern Washington; video is used by human experts to count fish by species. | Recorded in the Kvichak River in Alaska during the deployment of ORPC's RivGen instream turbines. |
| Species | fish, unknown species | Spring and summer run Chinook, Jack Chinook, Sockeye | salmon, adult and smolt |
| Video Image Size | 720 x 576 | 1280 x 960 | 320 x 240 |
| Frame Rate | 15 fps | 30 fps | 10 fps |
| Color Type | grayscale | color | color |
| Total Frames | 15918 | 24000 | 30277 |
| Frames with Fish | 7663 (48%) | 13405 (55%) | 1002 (3%) |

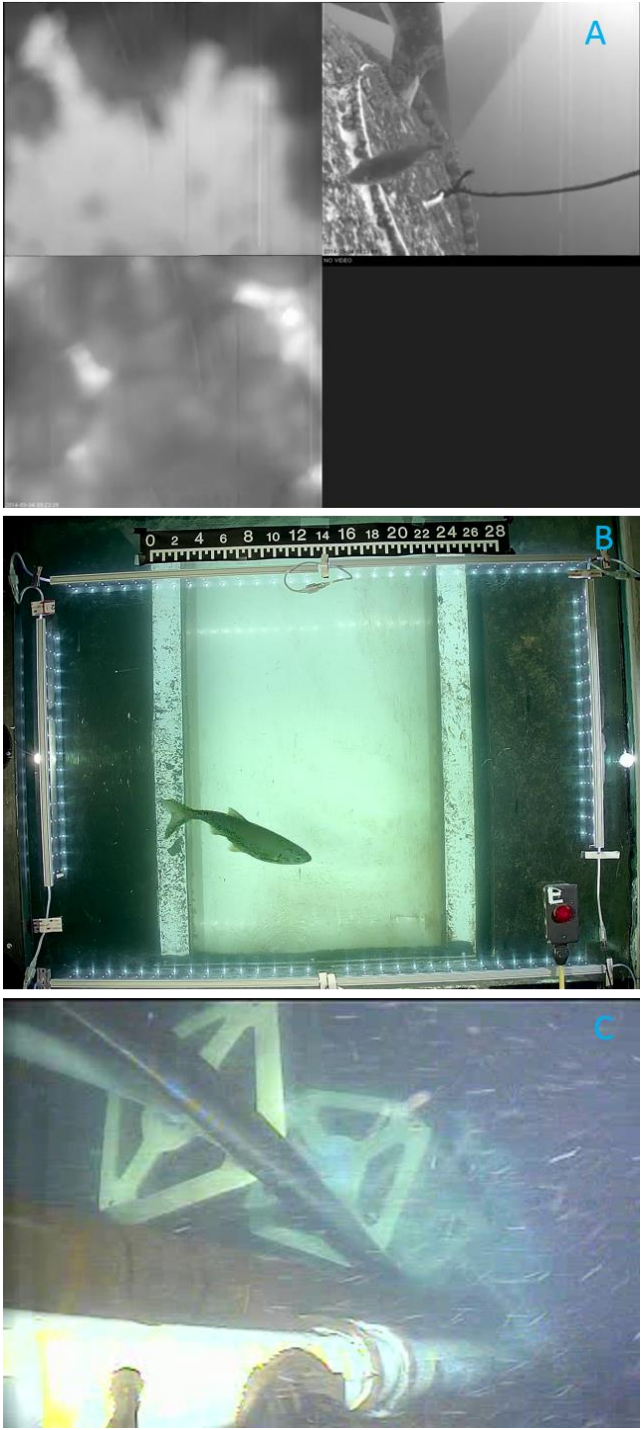

Fig. 1   Dataset examples: Voith Hydro (A), Wells Dam (B), and Igiugig (C)

Through the cooperation of public and private organizations who would benefit from automated processing, we have obtained three video datasets (Table I and Fig. 1). The Voith Hydro dataset was recorded at the Fall of Warness test site of the European Marine Energy Center in Scotland during a test of a tidal turbine prototype. The Wells Dam data were recorded at a hydroelectric dam in Washington State where such data is regularly recorded to monitor fish passage through the dam. The video is analyzed by human experts to count the fish. The Igiugig dataset was recorded in Alaska as part of a pilot demonstration of Ocean Renewable Power Company's RivGen instream turbine.  Selected video from each dataset were annotated using a script developed for that purpose.

## III. DEEP LEARNING APPROACH

### A. YOLO

Recent years have witnessed deep neural networks becoming the leading method for high quality computer vision, including identification, classification, and object detection. Many of the early successes of deep learning for object detections adapted the classification for detection purposes, in which the algorithm examine each of the proposed region of the image and conduct a separate classification. Faster R-CNN [18] is an example of this regional proposal method. On the other side, YOLO [14] treats object detection as a regression problem and output bounding box coordinates and classes confidence score directly. YOLO's architecture represents a more straightforward method for object detection comparing to regional proposal method, and also use much less computational time.

Hence, for our study of detecting fish from underwater videos near water power applications, we implemented version 3 of YOLO, using the Tensorflow and Keras implementation by Huynh [19]. We created a fork of Huynh's repository, and modified the data processing pipeline to train the YOLO v3 model on underwater fish videos converted images. We modified the code to allowing output of ground truth boxes in green and predicted boxes in red for both validation and testing samples. We also modified the code to output additional statistics in csv format, including false positives, true positive, precision, and recall.

### B. Transfer Learning

Transfer learning is a method in which a deep learning model developed for one task is reused as the parameter initialization for another model on a different task. We applied transfer learning using YOLO v3 [20]'s weights to initialize our training. YOLO v3 has 106 convolutional layers, and can detection a variety of different classes ranging from cats to cars but does not detect fish.

Our underwater fish videos were converted to ".PNG" format images for each frame to ease the process of batch loading. Annotation labels were converted to ".xml" format for each frame as well. During training, only images with fish detection were loaded to train the model, as a way to reduce training time. The model is trained using stochastic gradient descent optimization with Adam optimizer with learning rate of 0.0001. All training and testing were performed on a Windows workstation with a single Nvidia GEFORCE GTX 1080 Ti GPU.

## IV. EXPERIMENTS AND RESULTS

### A. Experiment I: Train and Evaluate on All Datasets

Our first experiment evaluates the general accuracy that YOLO can achieve using all three datasets. For each of the three fish video datasets, we divided them into roughly 80% - 20% for training and testing. We combined all training from three source as our training set, consisting of 54516 frames and associated annotations (Table II). From the testing set, we selected a subset of 10% of the testing set for validating purpose during training. We setup the model to train for 100 epochs, and stopped early at 20 epochs as the model starts to overfit. After the model was trained, we evaluated the fish detection accuracy on the testing set. Fig. 2 shows some examples of correct detection and incorrect detections, and Fig. 3 shows the precision recall curve for the testing sets.

We used the metric of mean Average Precision (mAP), which is the average of the maximum precisions at different recall values, to evaluate the model performance. When tested on individual source of dataset, the model achieved best results on Wells Dam dataset with 0.5575 mAP (Table 2). Wells Dam dataset's better performance than the other two datasets were as expected, considering its high resolution and three full color channels. However, the mAP metric on Wells Dam dataset is lower than our expectation. Fig. 2 F reveals that most of the false detection in Wells Dam dataset are at the side of fish ladder window, where only partial of the fish bodies are visible and difficult for even human annotators to correctly label. These hard cases are sometimes more numerous than fish passing through the white plate, which greatly reduced the overall mAP score on the Wells Dam dataset.

The model achieved a mAP of 0.5474 on the Voith Hydro dataset. The challenges with the Voith Hydro dataset are low resolution and being greyscale. Low resolution makes the single-frame detection difficult, as most fish appeared in this dataset does not contain detailed fish features. Greyscale makes the fish hard to be distinguished from the background. The model achieved satisfactory results in detecting most of the fish when a school of fish passed the ocean energy devices (Fig. 2 B). The model has roughly the same amount of false positives and true negatives. Since many fish in the Voith Hydro dataset are almost impossible to detect by human eyes when only presented with a single frame, we believe that training a network that leverages multiple frames can further improve the detection accuracy. The Voith Hydro dataset has a more gentle slopped precision-recall curve than the Wells Dam dataset (Fig. 3), indicating that the detection difficulty among objects in the Voith Hydro dataset is more homogeneous.

The model performed the worst on the Igiugig dataset, with a mAP of 0.4507. The water flows fast in this dataset, with air bubbles and debris passing by often of similar size to small fish. Also, only 3% of the frames contains fish, although the total number of frames is large, the actually number of frames that entered training is small (~1k) comparing to the other two datasets. The model did a good job in distinguishing bubbles and debris from fish, with very few false detections on these objects. The precision-recall curve dropped sharply after 50% recall, with precision dropping from 0.6 to 0.2. One of the factor that drove down the mAP is the false positives as show in the middle of Fig. 2 H, where the energy device shifted angles and a portion of the device were confused to be a fish by the model, and the error continued for the entire testing video. If we move the troublesome video to training set, the model would have learnt to remove the false positives, and generate better evaluation scores. However, our focus has been testing the model on real world cases rather than improving predictions scores. This interesting observation does point out the importance of including videos from different camera angles and lighting conditions, which will make the trained model more robust. In this application we did not train the model on frames without fish, however training the model on frames without fish can improve the model generalization because empty frames may contains useful information about the background environment.

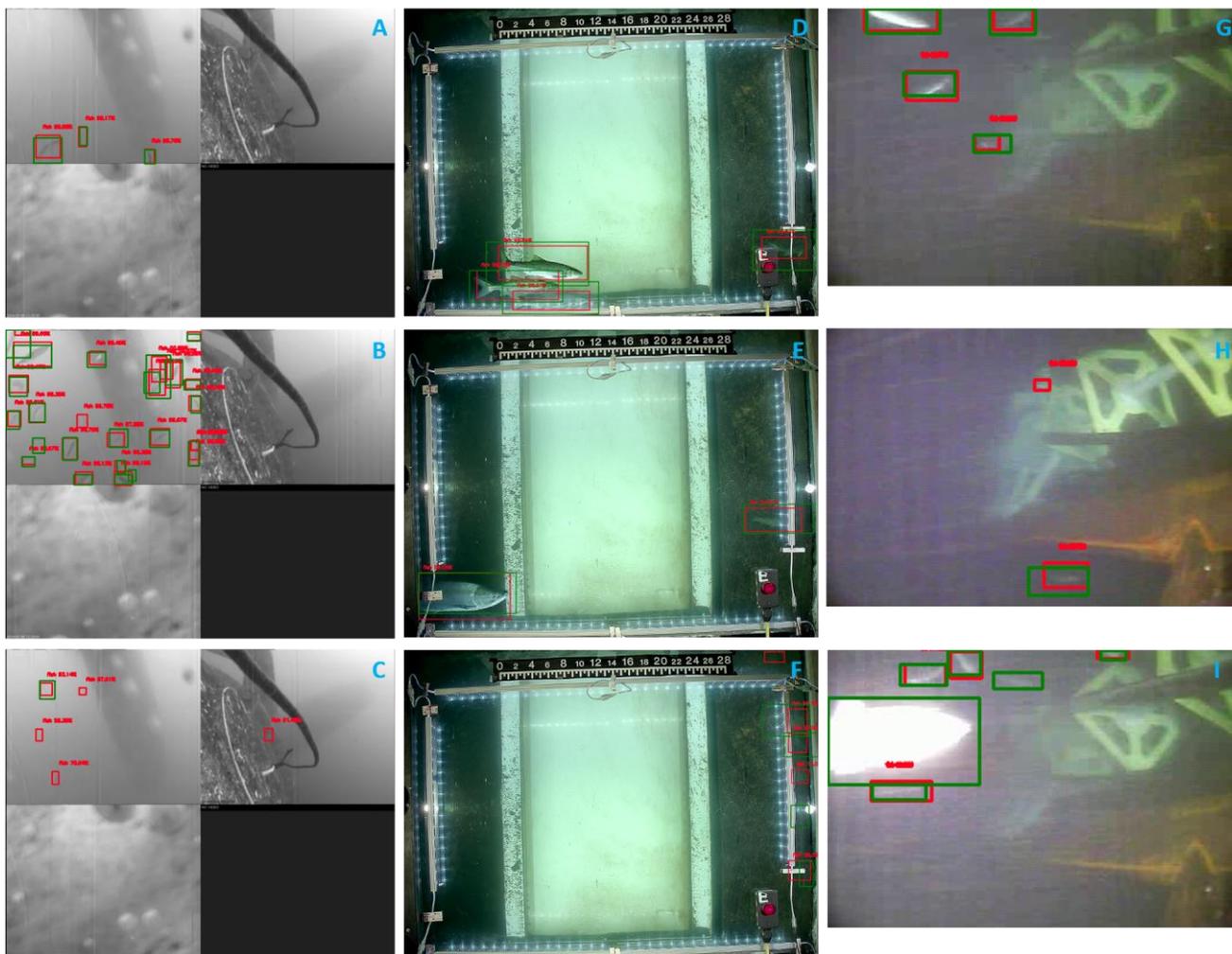

Fig. 2    Some correct detections and false detections in Experiment I. The green boxes are human annotations of fish, while the red boxes are model predictions along with labeled confidence score. One can tell why the detection task can be challenging even for human beings. A, B, C, G, H, and I show blurry images with small fish objects that are hardly distinguishable from environment. B shows school of fish, in which although the model correctly detected most of the fish, the model had some wrong detection boxes at overlapping fish. D and E show correct detections for Wells Dam dataset. F shows the challenges with Wells Dam dataset are often at the side of fish ladder window where only partial of the fish bodies are visible, causing the model to put detection boxes different from human annotators. I shows turbulent underwater environment, with an adult salmon reflecting flashlight and was missed by the model.

TABLE II. EXPERIMENT I DESIGN AND EVALUATION RESULTS

| Dataset | # of training frame / percent total training | # of testing frame / percent total testing | Testing mAP |
|---|---|---|---|
| Voith Hydro | 12819 / 23.5% | 3099 / 19.8% | 0.5474 |
| Wells Dam | 19200 / 35.2% | 4800 / 30.6% | 0.5575 |
| Igiugig | 22497 / 41.3% | 7780 / 49.6% | 0.4507 |
| Total | 54516 | 15679 | 0.5392 |

to be trained on the new dataset to be able to correctly detect fish.

We then continued the training with only the Igiugig dataset for 20 epochs, and achieved a mAP score of 0.3856 when tested on Igiugig. This score, despite showing improvements comparing to prior training score, is still lower than the score in Experiment I (0.4507). This result is the opposite of the Wells Dam and Voith Hydro datasets, where the scores are better when the model is only trained on a single dataset instead of a blend of datasets. We believe the reason lies in the internal variations of the datasets. The Igiugig datasets contain many camera angles and different environmental settings, and in the testing set there are many frames very different from the training set. Hence, when the model over-fits to the Igiugig training set, the testing result is worse than the model trained on all datasets, where the model is more robust. This phenomenon highlighted the importance of improving the diversity of underwater video training data to produce more generalized models, which is the value that our contributed datasets bring.

TABLE III. EXPERIMENT II DESIGN AND EVALUATION RESULTS

| Dataset | # of training frame | # of testing frame | Testing mAP |
|---|---|---|---|
| Voith Hydro | 12819 | 3099 | 0.5714 |
| Wells Dam | 19200 | 4800 | 0.5659 |
| Igiugig | 0 | 7780 | 0.0055 |

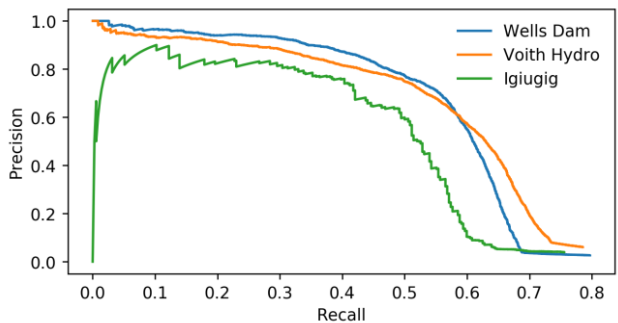

Fig. 3  The precision recall curve on testing sets in Experiment I.

### B. Experiment II: Reserve Igiugig Dataset to Test Generalization

Despite the success of deep learning in many computer vision challenges in recent years, the applications of pre-trained models on a new data source is rarely successful, due to overfitting problem and unlearned features that often exist in new data source. To test whether our model is robust enough on another data source, we conducted Experiment II. Just like Experiment I, we initiated the parameters with YOLOv3 weights. However, for training in Experiment II, we only used the Wells Dam and the Voith Hydro datasets, and did not show the Igiugig dataset to the model. The experiment design is described in Table III. Our testing results show that our trained model is indeed not generalized enough to correctly label Igiugig dataset if not trained on Igiugig data, with testing mAP score as 0.0055 on the Igiugig dataset (Table III, and Fig. 4). This result indicates when the model is applied to a different video dataset, the model need

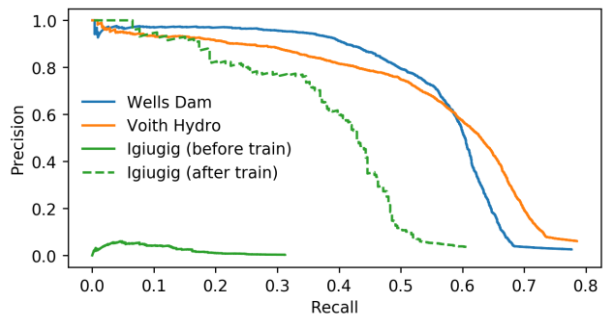

Fig 4.  The precision recall curve on testing sets in Experiment II. Before being trained on the Igiugig dataset, the model was only able to detect 30% of the fish with low precision (green line). After being trained, the model start to correctly detect fish in the Igiugig Dataset (green dashed line), and eventually achieved a mAP of 0.3856. This score, however, is still lower than the testing score in Experiment I.

## V. CONCLUSION AND FUTURE WORK

Automated fish detection in underwater environment is a challenging task, especially around the MHK and hydropower projects where the real world environment pose new difficulties including light, turbidity, high flow speed, etc. In this paper, we contributed frame-wise manually annotated underwater fish videos from three different MHK energy projects, adding to the currently small marine fish datasets which are mostly coral fish species. We evaluated YOLO, a state-of-art object detection model, using our annotated datasets. The results show that the detection accuracy is often limited by the video quality, indicating the need to install high quality cameras for future underwater energy projects to assist the ecological assessment task.

Deep learning has had many impressive successes reported in the literature for recognizing objects in still images, including individual video frames. Much of that work has been done using benchmark datasets such as COCO or ImageNet. While these datasets provide a wide variety of objects and image backgrounds, they are more representative of images found online than of images recorded in a real-world surveillance or monitoring study. The video data used for this project were recorded as part of ocean energy pilot studies and an hydroelectric dam monitoring program and fish are the only object class of interest. The COCO dataset does not contain a fish object class, so the usefulness of the pre-trained weights used for this study is questionable. ImageNet does contain a fish class with several species-specific subclasses, and so a model pre-trained on ImageNet may be more effective for detecting fish in these datasets. Due to differences in the video recordings -- background, aspect angle of fish, size of fish, resolution and frame rate -- we found that a model trained on two different datasets was not effective at recognizing fish in a completely new dataset. The results also showed that the model performed better on two of the datasets when it was not trained on the third.

There are several approaches we plan to try to further improve the underwater fish detection accuracy. First, image augmentation techniques, like cropping, flipping, rotating, and color shifting, may help to make the model more generalized and robust. Second, an ensemble of detection models may outperform individual models. Third, fish motion, which can be extracted from multi-frame ensembles, may be helpful to detect fish that appears vague in the videos. Such method has been implemented in detecting small flying bird near wind turbines [21] and can be tested in underwater videos.

ACKNOWLEDGMENT

The authors would like to thank Ocean Renewable Power Company, Aquatera and Douglas County Public Utility District for providing underwater video data to support this project. The authors would also like to thank PNNL interns Elvin Munoz, John Sederburg, Ann Archer, Sean McGaughey and Faigda Rico for annotating the video.

This work was funded by the U.S. Dept. of Energy's Energy Efficiency and Renewable Energy Technology Commercialization Fund.